# Gaussian Process Structural Equation Models with Latent Variables

**Ricardo Silva**
Department of Statistical Science
University College London
ricardo@stats.ucl.ac.uk

**Robert B. Gramacy**
Statistical Laboratory
University of Cambridge
bobby@statslab.cam.ac.uk

## Abstract

In a variety of disciplines such as social sciences, psychology, medicine and economics, the recorded data are considered to be noisy measurements of latent variables connected by some causal structure. This corresponds to a family of graphical models known as the structural equation model with latent variables. While linear non-Gaussian variants have been well-studied, inference in nonparametric structural equation models is still underdeveloped. We introduce a sparse Gaussian process parameterization that defines a non-linear structure connecting latent variables, unlike common formulations of Gaussian process latent variable models. The sparse parameterization is given a full Bayesian treatment without compromising Markov chain Monte Carlo efficiency. We compare the stability of the sampling procedure and the predictive ability of the model against the current practice.

## 1 CONTRIBUTION

A cornerstone principle of many disciplines is that observations are noisy measurements of hidden variables of interest. This is particularly prominent in fields such as social sciences, psychology, marketing and medicine. For instance, data can come in the form of social and economical indicators, answers to questionnaires in a medical exam or marketing survey, and instrument readings such as fMRI scans. Such indicators are treated as measures of latent factors such as the latent ability levels of a subject in a psychological study, or the abstract level of democratization of a country. The literature on structural equation models (SEMs) (Bartholomew et al., 2008; Bollen, 1989) approaches such problems with directed graphical models, where each node in the graph is a noisy function of its parents. The goals of the analysis include typical applications of latent variable models, such as projecting points in a latent space (with confidence regions) for ranking, clustering and visualization; density estimation; missing data imputation; and causal inference (Pearl, 2000; Spirtes et al., 2000).

This paper introduces a nonparametric formulation of SEMs with hidden nodes, where functions connecting latent variables are given a Gaussian process prior. An efficient but flexible sparse formulation is adopted. To the best of our knowledge, our contribution is the first full Gaussian process treatment of SEMs with latent variables.

We assume that the model graphical structure is given. Structural model selection with latent variables is a complex topic which we will not pursue here: a detailed discussion of model selection is left as future work. Asparouhov and Muthén (2009) and Silva et al. (2006) discuss relevant issues. Our goal is to be able to generate posterior distributions over parameters and latent variables with scalable sampling procedures with good mixing properties, while being competitive against non-sparse Gaussian process models.

In Section 2, we specify the likelihood function for our structural equation models and its implications. In Section 3, we elaborate on priors, Bayesian learning, and a sparse variation of the basic model which is able to handle larger datasets. Section 4 describes a Markov chain Monte Carlo (MCMC) procedure. Section 5 evaluates the usefulness of the model and the stability of the sampler in a set of real-world SEM applications with comparisons to modern alternatives. Finally, in Section 6 we discuss related work.

## 2 THE MODEL: LIKELIHOOD

Let $\mathcal{G}$ be a given directed acyclic graph (DAG). For simplicity, in this paper we assume that no observed variable is a parent in $\mathcal{G}$ of any latent variable. Many SEM applications are of this type (Bollen, 1989; Silva et al., 2006), and this will simplify our presentation. Likewise, we will treat models for continuous variables only. Although cyclic SEMs are also well-defined for the linear case (Bollen, 1989), non-linear cyclic models are not trivial to define and as such we will exclude them from this paper.

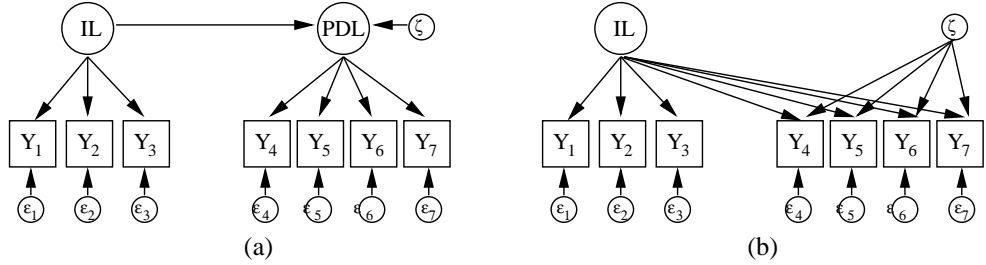

Figure 1: (a) An example adapted from Palomo et al. (2007): latent variable $IL$ corresponds to a scalar labeled as the industrialization level of a country. $PDL$ is the corresponding political democratization level. Variables $Y_1, Y_2, Y_3$ are indicators of industrialization (e.g., gross national product) while $Y_4, \ldots, Y_7$ are indicators of democratization (e.g., expert assessements of freedom of press). Each variable is a function of its parents with a corresponding additive error term: $\epsilon_i$ for each $Y_i$, and $\zeta$ for democratization levels. For instance, $PDL = f(IL) + \zeta$ for some function $f(\cdot)$. (b) Dependence among latent variables is essential to obtain sparsity in the measurement structure. Here we depict how the graphical dependence structure would look like if we regressed the observed variables on the independent latent variables of (a).

Let $\mathcal{X}$ be our set of latent variables and $X_i \in \mathcal{X}$ be a particular latent variable. Let $\mathbf{X}_{P_i}$ be the set of parents of $X_i$ in $\mathcal{G}$. The latent structure in our SEM is given by the following generative model: if the parent set of $X_i$ is not empty,

$$X_i = f_i(\mathbf{X}_{P_i}) + \zeta_i, \text{ where } \zeta_i \sim \mathcal{N}(0, v_{\zeta_i}) \quad (1)$$

$\mathcal{N}(m, v)$ is the Gaussian distribution with mean $m$ and variance $v$. If $X_i$ has no parents (i.e., it is an **exogenous** latent variable, in SEM terminology), it is given a mixture of Gaussians marginal[1].

The **measurement model**, i.e., the model that describes the distribution of observations $\mathcal{Y}$ given latent variables $\mathcal{X}$, is as follows. For each $Y_j \in \mathcal{Y}$ with parent set $\mathbf{X}_{P_j}$, we have

$$Y_j = \lambda_{j0} + \mathbf{X}_{P_j}^\mathsf{T} \Lambda_j + \epsilon_j, \text{where } \epsilon_j \sim \mathcal{N}(0, v_{\epsilon_j}) \quad (2)$$

Error terms $\{\epsilon_j\}$ are assumed to be mutually independent and independent of all latent variables in $\mathcal{X}$. Moreover, $\Lambda_j$ is a vector of linear coefficients $\Lambda_j = [\lambda_{j1} \ldots \lambda_{j|\mathbf{X}_{P_j}|}]^\mathsf{T}$. Following SEM terminology, we say that $Y_j$ is an **indicator** of the latent variables in $\mathbf{X}_{P_j}$.

An example is shown in Figure 1(a). Following the notation of Bollen (1989), squares represent observed variables and circles, latent variables. SEMs are graphical models with an emphasis on sparse models where: 1. latent variables are dependent according to a directed graph model; 2. observed variables measure (i.e., are children of) very few latent variables. Although sparse latent variable models have been the object of study in machine learning and statistics (e.g., Wood et al. (2006); Zou et al. (2006)), not much has been done on exploring nonparametric models with dependent latent structure (a loosely related exception being dynamic systems, where filtering is the typical application). Figure 1(b) illustrates how modeling can be affected by discarding the structure among latents[2].

### 2.1 Identifiability Conditions

Latent variable models might be unidentifiable. In the context of Bayesian inference, this is less of a theoretical issue than a computational one: unidentifiable models might lead to poor mixing in MCMC, as discussed in Section 5. Moreover, in many applications, the latent embedding of the data points is of interest itself, or the latent regression functions are relevant for causal inference purposes. In such applications, an unidentifiable model is of limited interest. In this Section, we show how to derive sufficient conditions for identifiability.

Consider the case where a latent variable $X_i$ has at least three unique indicators $\mathcal{Y}_i \equiv \{Y_{i\alpha}, Y_{i\beta}, Y_{i\gamma}\}$, in the sense that no element in $\mathcal{Y}_i$ has any other parent in $\mathcal{G}$ but $X_i$. It is known that in this case (Bollen, 1989) the parameters of the structural equations for each element of $\mathcal{Y}_i$ are identifiable (i.e., the linear coefficients and the error term variance) up to a scale and sign of the latent variable. This can be resolved by setting the linear structural equation of (say) $Y_{i\alpha}$ to $Y_{i\alpha} = X_i + \epsilon_{i\alpha}$. The distribution of the error terms is then identifiable. The distribution of $X_i$ follows from a deconvolution between the observed distribution of an element of $\mathcal{Y}_i$ and the identified distribution of the error term.

---
[1]For simplicity of presentation, in this paper we adopt a finite mixture of Gaussians marginal for the exogenous variables. However, introducing a Dirichlet process mixture of Gaussians marginal is conceptually straightforward.

[2]Another consequence of modeling latent dependencies is reducing the number of parameters of the model: a SEM with a linear measurement model can be seen as a type of module network (Segal et al., 2005) where the observed children of a particular latent $X_i$ share the same nonlinearities propagated from $\mathbf{X}_{P_i}$: in the context of Figure 1, each indicator $Y_i \in \{Y_4, \ldots, Y_7\}$ has a conditional expected value of $\lambda_{i0} + \lambda_{i1} f_2(X_1)$ for a given $X_1$: function $f_2(\cdot)$ is shared among the indicators of $X_2$.

Identifiability of the joint of $\mathcal{X}$ can be resolved by multivariate deconvolution under extra assumptions. For instance, Masry (2003) describes one setup for the problem in the context of kernel density estimation (with known joint distribution of error terms, but unknown joint of $\mathcal{Y}$).

Assumptions for the identifiability of functions $f_i(\cdot)$, given the identifiability of the joint of $\mathcal{X}$, have been discussed in the literature of error-in-variables regression (Fan and Truong, 1993; Carroll et al., 2004). Error-in-variables regression is a special case of our problem, where $X_i$ is observed but $\mathbf{X}_{P_i}$ is not. However, since we have $Y_{i\alpha} = X_i + \epsilon_i$, this is equivalent to a error-in-variables regression $Y_{i\alpha} = f_i(\mathbf{X}_{P_i}) + \epsilon_{i\alpha} + \zeta_i$, where the compound error term $\epsilon_{i\alpha} + \zeta_i$ is still independent of $\mathbf{X}_{P_i}$.

It can be shown that such identifiability conditions can be exploited in order to identify causal directionality among latent variables under additional assumptions, as discussed by Hoyer et al. (2008a) for the fully observed case[3]. In our context, we focus on the implications of identifiabilty on MCMC (Section 5).

## 3 THE MODEL: PRIORS

Each $f_i(\cdot)$ can be given a Gaussian process prior (Rasmussen and Williams, 2006). In this case, we call this class of models the GPSEM-LV family, standing for Gaussian Process Structural Equation Model with Latent Variables. Models without latent variables and measurement models have been discussed by Friedman and Nachman (2000)[4].

### 3.1 Gaussian Process Prior and Notation

Let $X_i$ be an arbitrary latent variable in the graph, with latent parents $\mathbf{X}_{P_i}$. We will use $\mathbf{X}^{(d)}$ to represent the $d^{th}$ $\mathbf{X}$ sampled from the distribution of random vector $\mathbf{X}$, and $X_i^{(d)}$ indexes its $i^{th}$ component. For instance, $\mathbf{X}_{P_i}^{(d)}$ is the $d^{th}$ sample of the parents of $X_i$. A training set of size $N$ is represented as $\{\mathbf{Z}^{(1)}, \ldots, \mathbf{Z}^{(N)}\}$, where $\mathbf{Z}$ is the set of all variables. Lower case $\mathbf{x}$ represents fixed values of latent variables, and $\mathbf{x}^{1:N}$ represents a whole set $\{\mathbf{x}^{(1)}, \ldots, \mathbf{x}^{(N)}\}$.

---

[3] Notice that if the distribution of the error terms is non-Gaussian, identification is easier: we only need two unique indicators $Y_{i\alpha}$ and $Y_{i\beta}$: since $\epsilon_{i\alpha}, \epsilon_{i\beta}$ and $X_i$ are mutually independent, identification follows from known results derived in the literature of overcomplete independent component analysis (Hoyer et al., 2008b).

[4] To see how the Gaussian process networks of Friedman and Nachman (2000) are a special case of GPSEM-LV, imagine a model where each latent variable is measured without error. That is, each $X_i$ has at least one observed child $Y_i$ such that $Y_i = X_i$. The measurement model is still linear, but each structural equation among latent variables can be equivalently written in terms of the observed variables: i.e., $X_i = f_i(\mathbf{X}_{P_i}) + \zeta_i$ is equivalent to $Y_i = f_i(\mathbf{Y}_{P_i}) + \zeta_i$, as in Friedman and Nachman.

For each $\mathbf{x}_{P_i}$, the corresponding Gaussian process prior for function values $\mathbf{f}_i^{1:N} \equiv \{f_i^{(1)}, \ldots, f_i^{(N)}\}$ is

$$\mathbf{f}_i^{1:N} \mid \mathbf{x}_{P_i}^{1:N} \sim \mathcal{N}(0, \mathbf{K}_i)$$

where $\mathbf{K}_i$ is a $N \times N$ kernel matrix (Rasmussen and Williams, 2006), as determined by $\mathbf{x}_{P_i}^{1:N}$. Each corresponding $x_i^{(d)}$ is given by $f_i^{(d)} + \zeta_i^{(d)}$, as in Equation (1).

MCMC can be used to sample from the posterior distribution over latent variables and functions. However, each sampling step in this model costs $\mathcal{O}(N^3)$, making sampling very slow when $N$ is at the order of hundreds, and essentially undoable when $N$ is in the thousands. As an alternative, we introduce a multilayered representation adapted from the pseudo-inputs model of Snelson and Ghahramani (2006). The goal is to reduce the sampling cost down to $\mathcal{O}(M^2 N)$, $M < N$. $M$ can be chosen according to the available computational resources.

### 3.2 Pseudo-inputs Review

We briefly review the pseudo-inputs model (Snelson and Ghahramani, 2006) in our notation. As before, let $\mathbf{X}^{(d)}$ represent the $d^{th}$ data point for some $\mathbf{X}$. For a set $\mathbf{X}_i^{1:N} \equiv \{X_i^{(1)}, \ldots, X_i^{(N)}\}$ with corresponding parent set $\mathbf{X}_{P_i}^{1:N} \equiv \{\mathbf{X}_{P_i}^{(1)}, \ldots, \mathbf{X}_{P_i}^{(N)}\}$ and corresponding latent function values $\mathbf{f}_i^{1:N}$, we define a *pseudo-input* set $\bar{\mathbf{X}}_i^{1:M} \equiv \{\bar{\mathbf{X}}_i^{(1)}, \ldots, \bar{\mathbf{X}}_i^{(M)}\}$ such that

$$\mathbf{f}_i^{1:N} \mid \mathbf{x}_{P_i}^{1:N}, \bar{\mathbf{f}}_i, \bar{\mathbf{x}}_i^{1:M} \sim \mathcal{N}(\mathbf{K}_{i;NM} \mathbf{K}_{i;M}^{-1} \bar{\mathbf{f}}_i, \mathbf{V}_i)$$
$$\bar{\mathbf{f}}_i \mid \bar{\mathbf{x}}_i^{1:M} \sim \mathcal{N}(0, \mathbf{K}_{i;M}) \quad (3)$$

where $\mathbf{K}_{i;NM}$ is a $N \times M$ matrix with each $(j, k)$ element given by the kernel function $k_i(\mathbf{x}_{P_i}^{(j)}, \bar{\mathbf{x}}_i^{(k)})$. Similarly, $\mathbf{K}_{i;M}$ is a $M \times M$ matrix where element $(j, k)$ is $k_i(\bar{\mathbf{x}}_i^{(j)}, \bar{\mathbf{x}}_i^{(k)})$. It is important to notice that each pseudo-input $\bar{\mathbf{X}}_i^{(d)}$, $d = 1, \ldots, M$, has the same dimensionality as $\mathbf{X}_{P_i}$. The motivation for this is that $\bar{\mathbf{X}}_i$ works as an alternative training set, with the original prior predictive means and variances being recovered if $M = N$ and $\bar{\mathbf{X}}_i = \mathbf{X}_{P_i}$.

Let $\mathbf{k}_{i;dM}$ be the $d^{th}$ row of $\mathbf{K}_{i;NM}$. Matrix $\mathbf{V}_i$ is a diagonal matrix with entry $v_{i;dd}$ given by $v_{i;dd} = k_i(\mathbf{x}_{P_i}^{(d)}, \mathbf{x}_{P_i}^{(d)}) - \mathbf{k}_{i;dM}^\mathsf{T} \mathbf{K}_{i;M}^{-1} \mathbf{k}_{i;dM}$. This implies that all latent function values $\{f_i^{(1)}, \ldots, f_i^{(N)}\}$ are conditionally independent.

### 3.3 Pseudo-inputs: A Fully Bayesian Formulation

The density function implied by (3) replaces the standard Gaussian process prior. In the context of Snelson and Ghahramani (2006), input and output variables are observed, and as such Snelson and Ghahramani optimize $\bar{\mathbf{x}}_i^{1:M}$ by maximizing the marginal likelihood of the model.

This is practical but sometimes prone to overfitting, since pseudo-inputs are in fact free parameters, and the pseudo-inputs model is best seen as a variation of the Gaussian process prior rather than an approximation to it (Titsias, 2009).

In our setup, there is limited motivation to optimize the pseudo-inputs since the inputs themselves are random variables. For instance, we show in the next section that the cost of sampling pseudo-inputs is no greater than the cost of sampling latent variables, while avoiding cumbersome optimization techniques to choose pseudo-input values. Instead we put a prior on the pseudo-inputs and extend the sampling procedure. By conditioning on the data, a good placement for the pseudo-inputs can be learned, since $\mathbf{X}_{P_i}$ and $\bar{\mathbf{X}}_i^{(d)}$ are dependent in the posterior. Moreover, it naturally provides a protection against overfitting.

A simple choice of priors for pseudo-inputs is as follows: each pseudo-input $\bar{\mathbf{X}}_i^{(d)}$, $d = 1, \ldots, M$, is given a $\mathcal{N}(\mu_i^d, \Sigma_i^d)$ prior, independent of all other random variables. A partially informative (empirical) prior can be easily defined in the case where, for each $X_k$, we have the freedom of choosing a particular indicator $Y_q$ with fixed structural equation $Y_q = X_k + \epsilon_q$ (see Section 2.1), implying $E[X_k] = E[Y_q]$. This means if $X_k$ is a parent $X_i$, we set the respective entry in $\mu_i^d$ (recall $\mu_i^d$ is a vector with an entry for every parent of $X_i$) to the empirical mean of $Y_q$. Each prior covariance matrix $\Sigma_i^d$ is set to be diagonal with a common variance.

Alternatively, we would like to spread the pseudo-inputs a priori: other things being equal, pseudo-inputs that are too close to each can be wasteful given their limited number. One prior, inspired by space-filling designs from the experimental design literature (Santner et al., 2003), is

$$p(\bar{\mathbf{x}}_i^{1:M}) \propto \det(\mathbf{D}_i)$$

the determinant of a kernel matrix $\mathbf{D}_i$. We use a squared exponential covariance function with characteristic length scale of 0.1 (Rasmussen and Williams, 2006), and a "nugget" constant that adds $10^{-4}$ to each diagonal term. This prior has support over a $[-L, L]^{|\mathbf{X}_{P_i}|}$ hypercube. We set $L$ to be three times the largest standard deviation of observed variables in the training data. This is the pseudo-input prior we adopt in our experiments, where we center all observed variables at their empirical means.

### 3.4 Other Priors

We adopt standard priors for the parametric components of this model: independent Gaussians for each coefficient $\lambda_{ij}$, inverse gamma priors for the variances of the error terms and a Dirichlet prior for the distribution of the mixture indicators of the exogenous variables.

## 4 INFERENCE

We use a Metropolis-Hastings scheme to sample from our space of latent variables and parameters. Similarly to Gibbs sampling, we sample blocks of random variables while conditioning on the remaining variables. When the corresponding conditional distributions are canonical, we sample directly from them. Otherwise, we use mostly standard random walk proposals.

Conditioned on the latent variables, sampling the parameters of the measurement model is identical to the case of classical Bayesian linear regression. We omit the expressions for simplicity. The same can be said of the sampling scheme for the posterior variances of each $\zeta_i$. Sampling the mixture distribution parameters for the exogenous variables is also identical to the standard Bayesian case of Gaussian mixture models, and also omitted.

We describe the remaining stages of the sampler for the sparse model. The sampler for the model with full Gaussian process priors is simpler and analogous.

### 4.1 Sampling Latent Functions

In principle, one can analytically marginalize the pseudo-functions $\bar{\mathbf{f}}_i^{1:M}$. However, keeping an explicit sample of the pseudo-functions is advantageous when sampling latent variables $X_i^{(d)}$, $d = 1, \ldots, N$: for each child $X_c$ of $X_i$, only the corresponding factor for the conditional density of $\mathbf{f}_c^{(d)}$ needs to be computed (at a $\mathcal{O}(M)$ cost), since function values are independent given latent parents and pseudo-functions. This issue does not arise in the fully-observed case of Snelson and Ghahramani (2006), who do marginalize the pseudo-functions.

Pseudo-functions and functions $\{\bar{\mathbf{f}}_i^{1:M}, \mathbf{f}_i^{1:N}\}$ are jointly Gaussian given all other random variables and data. The conditional distribution of $\bar{\mathbf{f}}_i^{1:M}$ given everything, except itself *and* $\{f_i^{(1)}, \ldots, f_i^{(N)}\}$, is Gaussian with covariance matrix

$$\bar{\mathbf{S}}_i \equiv (\mathbf{K}_{i;M}^{-1} + \mathbf{K}_{i;M}^{-1} \mathbf{K}_{i;NM}^{\mathsf{T}} (\mathbf{V}_i^{-1} + \mathbf{I}/v_{\zeta_i}) \mathbf{K}_{i;NM} \mathbf{K}_{i;M}^{-1})^{-1}$$

where $\mathbf{V}_i$ is defined in Section 3.2 and $\mathbf{I}$ is a $M \times M$ identity matrix. The total cost of computing this matrix is $\mathcal{O}(NM^2 + M^3) = \mathcal{O}(NM^2)$. The corresponding mean is

$$\bar{\mathbf{S}}_i \times \mathbf{K}_{i;M}^{-1} \mathbf{K}_{i;NM}^{\mathsf{T}} (\mathbf{V}_i^{-1} + \mathbf{I}/v_{\zeta_i}) \mathbf{x}_i^{1:N}$$

where $\mathbf{x}_i^{1:N}$ is a column vector of length $N$.

Given that $\bar{\mathbf{f}}_i^{1:M}$ is sampled according to this multivariate Gaussian, we can now sample $\{f_i^{(1)}, \ldots, f_i^{(N)}\}$ in parallel, since this becomes a mutually independent set with univariate Gaussian marginals. The conditional variance of $f_i^{(d)}$ is $v_i' \equiv 1/(1/v_{i;dd} + 1/v_{\zeta_i})$, where $v_{i;dd}$ is defined in Section

3.2. The corresponding mean is $v'_i(f_\mu^{(d)}/v_{i;dd} + x_i^{(d)}/v_{\zeta_i})$, where $f_\mu^{(d)} = \mathbf{k}_{i;dM}\mathbf{K}_{i;M}^{-1}\bar{\mathbf{f}}_i$.

In Section 5, we also sample from the posterior distribution of the hyperparameters $\Theta_i$ of the kernel function used by $\mathbf{K}_{i;M}$ and $\mathbf{K}_{i;NM}$. Plain Metropolis-Hastings is used to sample these hyperparameters, using an uniform proposal in $[\alpha\Theta_i, (1/\alpha)\Theta_i]$ for $0 < \alpha < 1$.

### 4.2 Sampling Pseudo-inputs and Latent Variables

We sample each pseudo-input $\bar{\mathbf{x}}_i^{(d)}$ one at a time, $d = 1, 2, \ldots, M$. Recall that $\bar{\mathbf{x}}_i^{(d)}$ is a vector, with as many entries as the number of parents of $X_i$. In our implementation, we propose all entries of the new $\bar{\mathbf{x}}_i^{(d)'}$ simultaneously using a Gaussian random walk proposal centered at $\bar{\mathbf{x}}_i^{(d)'}$ with the same variance in each dimension and no correlation structure. For problems where the number of parents of $X_i$ is larger than in our examples (i.e., four or more parents), other proposals might be justified.

Let $\bar{\pi}_i^{(\backslash d)}(\bar{\mathbf{x}}_i^{(d)})$ be the conditional prior for $\bar{\mathbf{x}}_i^{(d)}$ given $\bar{\mathbf{x}}_i^{(\backslash d)}$, where $(\backslash d) \equiv \{1, 2, \ldots, d-1, d+1, \ldots, M\}$. Given a proposed $\bar{\mathbf{x}}_i^{(d)'}$, we accept the new value with probability $\min\left\{1, l_i(\bar{\mathbf{x}}_i^{(d)'})/l_i(\bar{\mathbf{x}}_i^{(d)})\right\}$ where

$$l_i(\bar{\mathbf{x}}_i^{(d)}) = \bar{\pi}_i^{(\backslash d)}(\bar{\mathbf{x}}_i^{(d)'}) \times p(\bar{f}_i^{(d)} \mid \bar{\mathbf{f}}_i^{(\backslash d)}, \bar{\mathbf{x}}_i)$$
$$\times \prod_{d=1}^N v_{i;dd}^{-1/2} e^{-(f_i^{(d)} - \mathbf{k}_{i;dM}\mathbf{K}_{i;M}^{-1}\bar{\mathbf{f}}_i)^2/(2v_{i;dd})}$$

and $p(\bar{f}_i^{(d)} \mid \bar{\mathbf{f}}_i^{(\backslash d)}, \bar{\mathbf{x}}_i)$ is the conditional density that follows from Equation (3). Row vector $\mathbf{k}_{i;dM}$ is the $d^{th}$ row of matrix $\mathbf{K}_{i;NM}$. Fast submatrix updates of $\mathbf{K}_{i;M}^{-1}$ and $\mathbf{K}_{i;NM}\mathbf{K}_{i;M}^{-1}$ are required in order to calculate $l_i(\cdot)$ at a $\mathcal{O}(NM)$ cost, which can be done by standard Cholesky updates (Seeger, 2004). The total cost is therefore $\mathcal{O}(NM^2)$ for a full sweep over all pseudo-inputs.

The conditional density $p(\bar{f}_i^{(d)} \mid \bar{\mathbf{f}}_i^{(\backslash d)}, \bar{\mathbf{x}}_i)$ is known to be sharply peaked for moderate sizes of $M$ (at the order of hundreds) (Titsias et al., 2009), which may cause mixing problems for the Markov chain. One way to mitigate this effect is to also propose a value $\bar{f}_i^{(d)'}$ jointly with $\bar{\mathbf{x}}_i^{(d)'}$, which is possible at no additional cost. We propose the pseudo-function using the conditional $p(\bar{f}_i^{(d)} \mid \bar{\mathbf{f}}_i^{(\backslash d)}, \bar{\mathbf{x}}_i)$. The Metropolis-Hastings acceptance probability for this variation is then simplified to $\min\left\{1, l_i^0(\bar{\mathbf{x}}_i^{(d)'})/l_i(\bar{\mathbf{x}}_i^{(d)})\right\}$, where

$$l_i^0(\bar{\mathbf{x}}_i^{(d)}) = \bar{\pi}_i^{(\backslash d)}(\bar{\mathbf{x}}_i^{(d)'})$$
$$\times \prod_{d=1}^N v_{i;dd}^{-1/2} e^{-(f_i^{(d)} - \mathbf{k}_{i;dM}\mathbf{K}_{i;M}^{-1}\bar{\mathbf{f}}_i)^2/(2v_{i;dd})}$$

Finally, consider the proposal for latent variables $X_i^{(d)}$. For each latent variable $X_i$, the set of latent variable instantiations $\{X_i^{(1)}, X_i^{(2)}, \ldots, X_i^{(N)}\}$ is mutually independent given the remaining variables. We propose each new latent variable value $x_i^{(d)'}$ in parallel, and accept or reject it based on a Gaussian random walk proposal centered at the current value $x_i^{(d)}$. We accept the move with probability $\min\left\{1, h_{X_i}(x_i^{(d)'})/h_{X_i}(x_i^{(d)})\right\}$ where, if $X_i$ is not an exogenous variable in the graph,

$$\begin{aligned}h_{X_i}(x_i^{(d)}) &= e^{-(x_i^{(d)} - f_i^{(d)})^2/(2v_{\zeta_i})}\\&\quad \times \prod_{X_c \in \mathbf{X}_{C_i}} p(f_c^{(d)} \mid \bar{\mathbf{f}}_c, \bar{\mathbf{x}}_c, x_i^{(d)})\\&\quad \times \prod_{Y_c \in \mathbf{Y}_{C_i}} p(y_c^{(d)} \mid \mathbf{x}_{P_c}^{(d)})\end{aligned}$$

where $\mathbf{X}_{C_i}$ is the set of latent children of $X_i$ in the graph, and $\mathbf{Y}_{C_i}$ is the corresponding set of observed children.

The conditional $p(f_c^{(d)} \mid \bar{\mathbf{f}}_c, \bar{\mathbf{x}}_c, x_i^{(d)})$, which follows from (3), is a non-linear function of $x_i^{(d)}$, but crucially does not depend on any $x_i^{(\cdot)}$ variable except point $d$. The evaluation of this factor costs $\mathcal{O}(M^2)$. As such, sampling all latent values for $X_i$ takes $\mathcal{O}(NM^2)$.

The case where $X_i$ is an exogenous variable is analogous, given that we also sample the mixture component indicators of such variables.

## 5 EXPERIMENTS

In this evaluation Section[5], we briefly illustrate the algorithm in a synthetic study, followed by an empirical evaluation on how identifiability matters in order to obtain an interpretable distribution of latent variables. We end this section with a study comparing the performance our model in predictive tasks against common alternatives[6].

### 5.1 An Illustrative Synthetic Study

We generated data from a model of two latent variables $(X_1, X_2)$ where $X_2 = 4X_1^2 + \zeta_2$, $Y_i = X_1 + \epsilon_i$ for $i = 1, 2, 3$ and $Y_i = X_2 + \epsilon_i$, for $i = 4, 5, 6$. $X_1$ and all error terms follow standard Gaussians. Given a sample of 150 points from this model, we set the structural equations for $Y_1$ and $Y_4$ to have a zero intercept and unit slope for identifiability purposes. Observed data for $Y_1$ against $Y_4$ is shown in Figure 2(a), which suggests a noisy quadratic relationship (plotted in 2(b), but unknown to the model). We run a GPSEM-LV model with 50 pseudo-inputs. The expected posterior value of each latent pair $\{X_1^{(d)}, X_2^{(d)}\}$ for $d = 1, \ldots, 150$ is plotted in Figure 2(c). It is clear that we were able to reproduce the original non-linear functional relationship given noisy data using a pseudo-inputs model.

---

[5]MATLAB code to run all of our experiments is available at http://www.homepages.ucl.ac.uk/~ucgtrbd/.

[6]Some implementation details: we used the squared exponential kernel function $k(\mathbf{x}_p, \mathbf{x}_q) = a\exp(-\frac{1}{2b}|\mathbf{x}_p - \mathbf{x}_q|^2) + 10^{-4}\delta_{pq}$, where $\delta_{pq} = 1$ is $p = q$ and 0 otherwise. The hyperprior for $a$ is a mixture of a gamma $(1, 20)$ and a gamma $(10, 10)$ with equal probability each. The same (independent) prior is given to $b$. Variance parameters were given inverse gamma $(2, 1)$ priors, and the linear coefficients were given Gaussian priors with a common large variance of 5. Exogenous latent variables were modeled as a mixture of five Gaussians where the mixture distribution is given a Dirichlet prior with parameter 10. Finally, for each latent $X_i$ variable we choose one of its indicators $Y_j$ and fix the corresponding edge coefficient to 1 and intercept to 0 to make the model identifiable. We perform 20,000 MCMC iterations with a burn-in period of 2000 (only 6000 iterations with 1000 of burn-in for the non-sparse GPSEM-LV due to its high computational cost). Small variations in the priors for coefficients (using a variance of 10) and variance parameters (using an inverse gamma $(2, 2)$), and a mixture of 3 Gaussians instead of 5, were attempted with no significant differences between models.

For comparison, the output of the Gaussian process latent variable model (GPLVM, Lawrence, 2005) with two hidden variables is shown in Figure 2(d). GPLVM here assumes that the marginal distribution of each latent variable is a standard Gaussian, but the measurement model is non-parametric. In theory, GPLVM is as flexible as GPSEM-LV in terms of representing observed joints. However, it does not learn functional relationships among latent variables, which is often of central interest in SEM applications (Bollen, 1989). Moreover, since no marginal dependence among latent variables is allowed, the model adapts itself to find (unidentifiable) functional relationships between the exogenous latent variables of the true model and the observables, analogous to the case illustrated by Figure 1(b). As a result, despite GPLVM being able to depict, as expected, some quadratic relationship (up to a rotation), it is noisier than the one given by GPSEM-LV.

### 5.2 MCMC and Identifiability

We now explore the effect of enforcing identifiability constraints on the MCMC procedure. We consider the dataset **Consumer**, a study[7] with 333 university students in Greece (Bartholomew et al., 2008). The aim of the study was to identify the factors that affect willingness to pay more to consume environmentally friendly products. We selected 16 indicators of environmental beliefs and attitudes, measuring a total of 4 hidden variables. For simplicity, we will call these variables $X_1, \ldots, X_4$. The structure among latents is $X_1 \to X_2$, $X_1 \to X_3$, $X_2 \to X_3$, $X_2 \to X_4$. Full details are given by Bartholomew et al. (2008).

All observed variables have a single latent parent in the corresponding DAG. As discussed in Section 2.1, the corresponding measurement model is identifiable by fixing the structural equation for one indicator of each variable to have a zero intercept and unit slope (Bartholomew et al., 2008). If the assumptions described in the references of Section 2.1 hold, then the latent functions are also identifiable. We normalized the dataset before running the MCMC

---

[7]There was one latent variable marginally independent of everything else. We eliminated it and its two indicators, as well as the REC latent variable that had only 1 indicator.

inference algorithm.

An evaluation of the MCMC procedure is done by running and comparing 5 independent chains, each starting from a different point. Following Lee (2007), we evaluate convergence using the EPSR statistic (Gelman and Rubin, 1992), which compares the variability of a given marginal posterior within each chain and between chains. We calculate this statistic for all latent variables $\{X_1, X_2, X_3, X_4\}$ across all 333 data points.

A comparison is done against a variant of the model where the measurement model is *not* sparse: instead, each observed variable has all latent variables as parents, and no coefficients are fixed. The differences are noticeable and illustrated in Figure 3. Box-plots of EPSR for the 4 latent variables are shown in Figure 4. It is difficult to interpret or trust an embedding that is strongly dependent on the initialization procedure, as it is the case for the unidentifiable model. As discussed by Palomo et al. (2007), identifiability might not be a fundamental issue for Bayesian inference, but it is an important practical issue in SEMs.

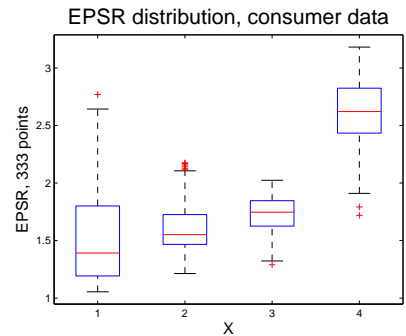

Figure 4: Boxplots for the EPSR distribution across each of the 333 datapoints of each latent variable. Boxes represent the distribution for the non-sparse model. A value less than 1.1 is considered acceptable evidence of convergence (Lee, 2007), but this essentially never happens. For the sparse model, all EPSR statistics were under 1.03.

### 5.3 Predictive Verification of the Sparse Model

We evaluate how well the sparse GPSEM-LV model performs compared against two parametric SEMs and GPLVM. The *linear* structural equation model is the SEM, where each latent variable is given by a linear combination of its parents with additive Gaussian noise. Latent variables without parents are given the same mixture of Gaussians model as our GPSEM-LV implementation. The *quadratic* model includes all quadratic and linear terms, plus first-order interactions, among the parents of any given latent variable. This is perhaps the most common non-linear SEM used in practice (Bollen and Paxton, 1998; Lee, 2007). GPLVM is fit with 50 active points and the rbf kernel with automatic relevance determination (Lawrence, 2005). Each sparse GPSEM model uses 50 pseudo-points.

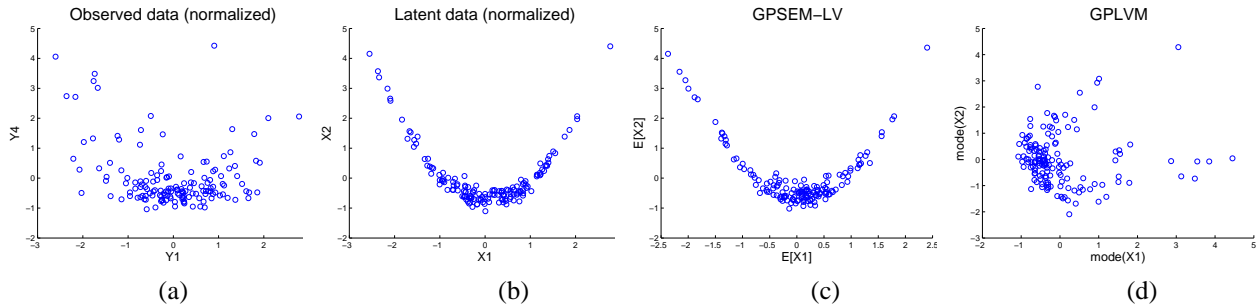

Figure 2: (a) Plot of observed variables $Y_1$ and $Y_4$ generated by adding standard Gaussian noise to two latent variables $X_1$ and $X_2$, where $X_2 = 4X_1^2 + \zeta_2$, $\zeta_2$ also a standard Gaussian. 150 data points were generated. (b) Plot of the corresponding latent variables, which are not recorded in the data. (c) The posterior expected values of the 150 latent variable pairs according to GPSEM-LV. (d) The posterior modes of the 150 pairs according to GPLVM.

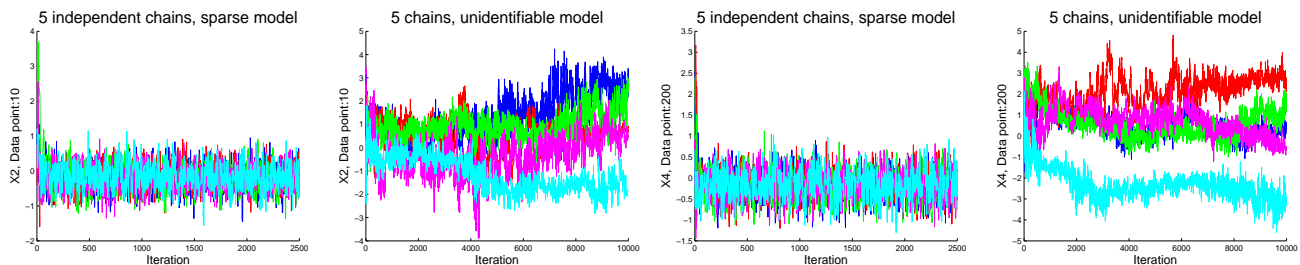

Figure 3: An illustration of the behavior of independent chains for $X_2^{(10)}$ and $X_4^{(200)}$ using two models for the **Consumer** data: the original (sparse) model (Bartholomew et al., 2008); an (unidentifiable) alternative where the each observed variable is an indicator of all latent variables. In the unidentifiable model, there is no clear pattern across the independent chains. Our model is robust to initialization, while the alternative unidentifiable approach cannot be easily interpreted.

We performed a 5-fold cross-validation study where the average predictive log-likelihood on the respective test sets is reported. Three datasets are used. The first is the **Consumer** dataset, described in the previous section.

The second is the **Abalone** data (Asuncion and Newman, 2007), where we postulate two latent variables, "Size" and "Weight." *Size* has as indicators the length, diameter and height of each abalone specimen, while *Weight* has as indicators the four weight variables. We direct the relationship among latent variables as $Size \rightarrow Weight$.

The third is the **Housing** dataset (Asuncion and Newman, 2007; Harrison and Rubinfeld, 1978), which includes indicators about features of suburbs in Boston that are relevant for the housing market. Following the original study (Harrison and Rubinfeld, 1978, Table IV), we postulate three latent variables: "Structural," corresponding to the structure of each residence; "Neighborhood," corresponding to an index of neighborhood attractiveness; and "Accessibility," corresponding to an index of accessibility within

Boston[8]. The corresponding 11 non-binary observed variables that are associated with the given latent concepts are used as indicators. The "Neighborhood" concept was refined into two, "Neighborhood I" and "Neighborhood II" due to the fact that three of its original indicators have very similar (and highly skewed) marginal distributions, which were very dissimilar from the others[9]. The structure among latent variables is given by a fully connected network directed according to the order {Accessibility, Structural, Neighborhood II, Neighborhood I}. Harrison and Rubinfeld (1978) provide full details on the meaning of the indicators. We note that it is well known that the **Hous-**

---

[8] The analysis by (Harrison and Rubinfeld, 1978, Table IV) also included a fourth latent concept of "Air pollution," which we removed due to the absence of one of its indicators in the electronic data file that is available.

[9] The final set of indicators, using the nomenclature of the UCI repository documentation file, is as follows: "Structural" has as indicators $RM$ and $AGE$; "Neighborhood I" has as indicators $CRIM$, $ZN$ and $B$; "Neighborhood II" has as indicators $INDUS, TAX, PTRATIO$ and $LSTAT$; "Accessibility" has as indicators $DIS$ and $RAD$. See (Asuncion and Newman, 2007) for detailed information about these indicators. Following Harrison and Rubinfeld, we log-transformed some of the variables: $INDUS, DIS, RAD$ and $TAX$.

**ing** dataset poses stability problems to density estimation due to discontinuities in the variable $RAD$, one of the indicators of accessibility (Friedman and Nachman, 2000). In order to get more stable results, we use a subset of the data (374 points) where $RAD < 24$.

The need for non-linear SEMs is well-illustrated by Figure 5, where fantasy samples of latent variables are generated from the predictive distributions of two models.

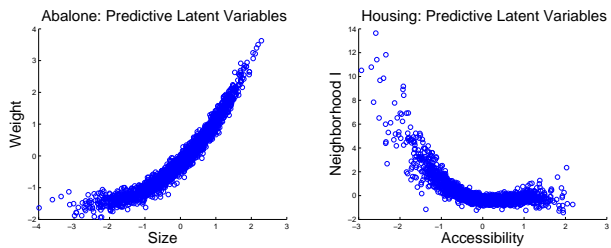

Figure 5: Scatterplots of 2000 fantasy samples taken from the predictive distributions of sparse GPSEM-LV models. In contrast, GPLVM would generate spherical Gaussians.

We also evaluate how the non-sparse GPSEM-LV behaves compared to the sparse alternative. Notice that while **Consumer** and **Housing** have each approximately 300 training points in each cross-validation fold, **Abalone** has over 3000 points. For the non-sparse GPSEM, we subsampled all of **Abalone** training folds down to 300 samples.

Results are presented in Table 1. Each dataset was chosen to represent a particular type of problem. The data in **Consumer** is highly linear. In particular, it is important to point out that the GPSEM-LV model is able to behave as a standard structural equation model if necessary, while the quadratic polynomial model shows some overfitting. The **Abalone** study is known for having clear functional relationships among variables, as also discussed by Friedman and Nachman (2000). In this case, there is a substantial difference between the non-linear models and the linear one, although GPLVM seems suboptimal in this scenario where observed variables can be easily clustered into groups. Finally, functional relationships among variables in **Housing** are not as clear (Friedman and Nachman, 2000), with multimodal residuals. GPSEM still shows an advantage, but all SEMs are suboptimal compared to GPLVM. One explanation is that the DAG on which the models rely is not adequate. Structure learning might be necessary to make the most out of nonparametric SEMs.

Although results suggest that the sparse model behaved better that the non-sparse one (which was true of some cases found by Snelson and Ghahramani, 2006, due to heteroscedasticity effects), such results should be interpreted with care. **Abalone** had to be subsampled in the non-sparse case. Mixing is harder in the non-sparse model since all datapoints $\{X_i^{(1)}, \ldots, X_i^{(N)}\}$ are dependent. While we believe that with larger sample sizes and denser latent structures the non-sparse model should be the best, large sample sizes are too expensive to process and, in many SEM applications, latent variables have very few parents.

It is also important to emphasize that the wallclock sampling time for the non-sparse model was an order of magnitude larger than the sparse case with $M = 50$. The sparse pseudo-inputs model was faster even considering that 3000 training points were used by the sparse model in the **Abalone** experiment, against 300 points by the non-sparse alternative.

## 6 RELATED WORK

Non-linear factor analysis has been studied for decades in the psychometrics literature[10]. A review is provided by Yalcin and Amemiya (2001). However, most of the classic work is based on simple parametric models. A modern approach based on Gaussian processes is the Gaussian process latent variable model of Lawrence (2005). By construction, factor analysis cannot be used in applications where one is interested in learning functions relating latent variables, such as in causal inference. For embedding, factor analysis is easier to use and more robust to model misspecification than SEM analysis. Conversely, it does not benefit from well-specified structures and might be harder to interpret. Bollen (1989) discusses the interplay between factor analysis and SEM. Practical non-linear structural equation models are discussed by Lee (2007), but none of such approaches rely on nonparametric methods. Gaussian processes latent structures appear mostly in the context of dynamical systems (e.g., Ko and Fox (2009)). However, the connection is typically among data points only, not among variables within a data point, where on-line filtering is the target application.

## 7 CONCLUSION

The goal of graphical modeling is to exploit the structure of real-world problems, but the latent structure is often ignored. We introduced a new nonparametric approach for SEMs by extending a sparse Gaussian process prior as a fully Bayesian procedure. Although a standard MCMC algorithm worked reasonably well, it is possible as future work to study ways of improving mixing times. This can be particularly relevant in extensions to ordinal variables, where the sampling of thresholds will likely make mixing more difficult. Since the bottleneck of the procedure is the sampling of the pseudo-inputs, one might consider a hybrid approach where a subset of the pseudo-inputs is fixed and determined prior to sampling using a cheap heuristic. New ways of deciding pseudo-input locations based on a given measurement model will be required. Evaluation with larger datasets (at least a few hundred variables) remains an open problem. Finally, finding ways of determining the graphical structure is also a promising area of research.

---

[10]Another instance of the "whatever you do, somebody in psychometrics already did it long before" law: http://www.stat.columbia.edu/~cook/movabletype/archives/2009/01/a_longstanding.html

Table 1: Average predictive log-likelihood in a 5-fold cross-validation setup. The five methods are the GPSEM-LV model with 50 pseudo-inputs (GPS), GPSEM-LV with standard Gaussian process priors (GP), the linear and quadratic structural equation models (LIN and QDR) and the Gaussian process latent variable model (GPL) of Lawrence (2005), a nonparametric factor analysis model. For **Abalone**, GP uses a subsample of the training data. The p-values given by a paired Wilcoxon signed-rank test, measuring the significance of positive differences between sparse GPSEM-LV and the quadratic model, are 0.03 (for **Consumer**), 0.34 (**Abalone**) and 0.09 (**Housing**).

|  | Consumer | | | | | Abalone | | | | | Housing | | | | |
|---|---|---|---|---|---|---|---|---|---|---|---|---|---|---|---|
|  | GPS | GP | LIN | QDR | GPL | GPS | GP | LIN | QDR | GPL | GPS | GP | LIN | QDR | GPL |
| Fold 1 | -20.66 | -21.17 | -20.67 | -21.20 | -22.11 | -1.96 | -2.08 | -2.75 | -2.00 | -3.04 | -13.92 | -14.10 | -14.46 | -14.11 | -11.94 |
| Fold 2 | -21.03 | -21.15 | -21.06 | -21.08 | -22.22 | -1.90 | -2.97 | -2.52 | -1.92 | -3.41 | -15.07 | -17.70 | -16.20 | -15.12 | -12.98 |
| Fold 3 | -20.86 | -20.88 | -20.84 | -20.90 | -22.33 | -1.91 | -5.50 | -2.54 | -1.93 | -3.65 | -13.66 | -15.75 | -14.86 | -14.69 | -12.58 |
| Fold 4 | -20.79 | -21.09 | -20.78 | -20.93 | -22.03 | -1.77 | -2.96 | -2.30 | -1.80 | -3.40 | -13.30 | -15.98 | -14.05 | -13.90 | -12.84 |
| Fold 5 | -21.26 | -21.76 | -21.27 | -21.75 | -22.72 | -3.85 | -4.56 | -4.67 | -3.84 | -4.80 | -13.80 | -14.46 | -14.67 | -13.71 | -11.87 |

**Acknowledgements**

We thank Patrick Hoyer and Ed Snelson for several useful discussions, and Irini Moustaki for the consumer data.